\documentclass{article}
\usepackage{kr}

\usepackage{times}
\usepackage{soul}
\usepackage{url}
\usepackage[hidelinks]{hyperref}
\usepackage[utf8]{inputenc}
\usepackage[small]{caption}
\usepackage{graphicx}
\usepackage{amsmath}
\usepackage{amsthm}
\usepackage{booktabs}
\usepackage{multirow}
\usepackage{array}
\newtheorem{definition}{Definition}

\title{Multi-Agent Legal Verifier Systems for Data Transfer Planning}

\author{%
Ha Thanh Nguyen$^{1,2}$ \and
Wachara Fungwacharakorn$^1$\and
Ken Satoh$^1$\\
\affiliations
$^1$ Center of Juris-Informatics, Joint Support-Center for Data Science Research, ROIS, Tokyo, Japan\\
$^2$ Research and Development Center for Large Language Models, NII, ROIS, Tokyo, Japan
}

\begin{document}

\maketitle

\begin{abstract}
Legal compliance in AI-driven data transfer planning is becoming increasingly critical under stringent privacy regulations such as the Japanese Act on the Protection of Personal Information (APPI). We propose a multi-agent legal verifier that decomposes compliance checking into specialized agents for statutory interpretation, business context evaluation, and risk assessment, coordinated through a structured synthesis protocol. Evaluated on a stratified dataset of 200 Amended APPI Article 16 cases with clearly defined ground truth labels and multiple performance metrics, the system achieves 72\% accuracy, which is 21 percentage points higher than a single-agent baseline, including 90\% accuracy on clear compliance cases (vs. 16\% for the baseline) while maintaining perfect detection of clear violations. While challenges remain in ambiguous scenarios, these results show that domain specialization and coordinated reasoning can meaningfully improve legal AI performance, providing a scalable and regulation-aware framework for trustworthy and interpretable automated compliance verification.

\end{abstract}

\section{Introduction}

The proliferation of artificial intelligence systems in data-intensive applications has created an urgent need for automated legal compliance verification. As organizations increasingly rely on AI agents for data transfer planning and execution, ensuring adherence to privacy regulations has become a critical challenge. This requires embedding legal reasoning \cite{nguyen2025llms} as a core capability within AI decision-making processes. The Japanese Amended Act on the Protection of Personal Information (Amended APPI), particularly Article 16 concerning Purpose Limitation, exemplifies the complex regulatory landscape that AI systems must navigate.

Traditional approaches to legal compliance checking typically employ single-agent architectures that attempt to capture the full complexity of legal reasoning within a monolithic system. However, legal compliance involves multiple dimensions of analysis: legal interpretation, contextual understanding, risk assessment, and precedent consideration. This multifaceted nature suggests that specialized multi-agent approaches may offer superior performance.

This paper introduces a novel multi-agent legal verifier system designed specifically for data transfer planning scenarios. Our approach transforms high-level plans into lists of specific actions, each of which is systematically evaluated by specialized legal verifier agents. The core innovation lies in the decomposition of legal compliance checking into specialized roles: legal analysis, context evaluation, and risk assessment, coordinated by a decision-making agent.

We present a comprehensive experimental evaluation comparing single-agent and multi-agent approaches across 200 carefully constructed test cases representing diverse compliance scenarios under Amended APPI Article 16. Our results demonstrate significant performance improvements for the multi-agent system, particularly in complex compliance scenarios requiring nuanced legal reasoning.

The contributions of this work are threefold: (1) a novel multi-agent architecture for legal compliance verification in data transfer planning, (2) a comprehensive experimental framework for evaluating legal AI systems, and (3) empirical evidence demonstrating the superiority of multi-agent approaches for complex legal reasoning tasks.

The rest of this paper is organized as follows. Section \ref{sec:related-work} reviews prior work on applying knowledge representation, multi-agent systems, and subsymbolic AI to legal compliance. Section \ref{sec:problem} provides the formalization of data transfer planning, APPI Article 16 compliance, and the legal verification problem. Section \ref{sec:system} details the proposed system. Section \ref{sec:experiment} outlines the experimental design. Section \ref{sec:result} reports the results and the detailed analysis. Section \ref{sec:discussion} discusses our findings. Finally, Section \ref{sec:conclusion} concludes the paper.

\section{Related Work}
\label{sec:related-work}

Legal compliance has been explored across multiple areas of AI, including knowledge representation, multi-agent systems, and subsymbolic approaches. This section reviews previous work in each of these areas.

\subsection{Knowledge Representation and Legal Compliance}

Researchers have long sought to formalize legal compliance using various forms of logic. Because legal compliance inherently involves deontic concepts -- such as obligations, prohibitions, and permissions -- it is a natural application of deontic logic \cite{von1951deontic}. It also requires temporal logic \cite{alur1990model} to capture the initiation and termination of actions, regulations, and their effects. Moreover, real-world legal systems contain exceptions and conflicts, making defeasible logic \cite{nute1994defeasible} a key foundation for modeling legal compliance. To achieve more practical formalizations, researchers have also explored combinations of these logics, including defeasible deontic logic \cite{nute1997defeasible} and temporal defeasible logic \cite{governatori2007temporal}. 

Building on these logical foundations, several first-order knowledge representations have been developed to formalize legal compliance  \cite{robaldo2024compliance}. PROLEG \cite{satoh2010proleg} is a Prolog-based framework that models exceptions explicitly, rather than using negation as failure, to better simulate legal reasoning. LegalRuleML \cite{palmirani2011legalruleml} is an XML-based representation that supports conflict resolution through superiority relations. Institution Action Language \cite{padget2016inst} is a declarative language designed to represent institutional actions and deontic concepts. 

Recent research has focused on complex scenarios such as business process compliance. Modeling business processes often requires semantic annotations (e.g., “control tags” or “flow tags”) to link tasks with compliance requirements. The process typically involves generating execution traces, identifying triggered obligations, and checking for fulfillment, violations, or compensations \cite{sadiq2014managing}. Due to its declarative nature and suitability for automating compliance checking over complex scenarios and event traces, answer set programming (ASP) has been widely applied in this domain \cite{cliffe2006answer,arias2024automated}.

In addition, researchers have investigated methods for explaining legal compliance and facilitating information sharing. Argumentation is a prominent approach to explaining compliance decisions \cite{burgemeestre2011value,prakken2015law}. Several systematic verification methods for information sharing have also been explored \cite{amor2021promise}, though primarily in the context of building codes rather than data protection regulations. Nonetheless, the principles of automated verification in these domains offer valuable insights for the design of legal compliance systems.

\subsection{Multi-Agent Systems and Legal Compliance}

Researchers have extensively examined legal compliance in multi-agent systems, particularly focusing on decentralizing compliance checking. Steenhuisen et al. \shortcite{steenhuisen2006framework} demonstrated how multi-agent planning frameworks can manage complex coordination tasks among non-cooperative agents by formalizing task decomposition, modeling inter-agent dependencies, and analyzing coordination complexity --insights directly applicable to legal compliance. Alechina et al. \shortcite{alechina2016decentralised} introduced a decentralized norm-monitoring framework for open multi-agent environments, demonstrating how agents can effectively detect and address norm violations without relying on a centralized authority. Their approach leverages an incentive-based mechanism in which agents both perform their primary tasks and monitor each other’s behaviour, with monitoring costs covered through a scrip system. This design enables the achievement of perfect monitoring at equilibrium or the adjustment of monitoring frequency to maintain low violation probabilities, highlighting the importance of allocating specialized monitoring responsibilities across different compliance dimensions.

Other studies have explored legal compliance in different multi-agent contexts, such as multi-level governance and epistemic planning. King et al. \shortcite{king2017automated} examined multi-level governance compliance, designing institutional structures where higher-level institutions govern lower-level ones. Liu and Liu \shortcite{liu2018multi} extended epistemic planning frameworks to handle common knowledge via a novel normal form in KD45 logic, and implemented the MEPC planner capable of generating solutions in domains requiring coordination through shared knowledge. 

Recent advances in multi-agent systems have shown promising applications in combining legal and ethical compliance.  
Hayashi et al. \shortcite{hayashi2023multi} proposed a multi-agent online planning architecture for real-time compliance, combining legal and ethical compliance mechanisms for trustworthy AI. Their work demonstrated the feasibility of integrating multiple specialized agents for regulatory compliance, though focused primarily on planning rather than verification.

%

\subsection{Subsymbolic AI and Legal Compliance}

With the rapid advancement of subsymbolic artificial intelligence, its implications for legal compliance have attracted growing attention. Mitrou \shortcite{mitrou2018data} offered a comprehensive analysis of the relationship between the GDPR and AI, especially subsymbolic AI, questioning whether current data protection regulations are truly “AI-proof.” This work emphasized the need for specialized systems capable of interpreting complex regulatory requirements in AI contexts. Javed and Li \shortcite{javed2024artificial} investigated semantic bias classification in legal judgments using subsymoblic AI, illustrating the potential of automated legal analysis while also identifying challenges in ensuring fairness and accuracy. Cerqueira \shortcite{cerqueira2024trust} examined the trustworthiness of LLM-based multi-agent systems for ethical AI, including GDPR compliance and fairness evaluation, highlighting both the promise and the limitations of current LLM approaches for legal reasoning.

To leverage the strengths and mitigate the weaknesses of both symbolic and subsymbolic AI, some researchers have proposed hybrid approaches. Nguyen et al. \cite{nguyen2023beyond} extended the PROLEG logic-programming framework for Japanese legal reasoning by integrating deep learning techniques -- such as fact extraction and end-to-end text interpretation -- to enhance interpretability, feasibility, and alignment with practitioners’ needs in compliance-checking systems. Nguyen and Satoh \shortcite{thanh2024krag} also introduced the KRAG (Knowledge Representation Augmented Generation) framework, which incorporates inference graphs via Soft PROLEG to guide LLMs in structured legal reasoning, thereby improving both precision and explainability in compliance-related outputs.

\section{Problem Formulation}
\label{sec:problem}

This section provides the formalization of data transfer planning, the Japanese Amended Act on the Protection of Personal Information (Amended APPI) Article 16 compliance, and legal verfier problem.

\subsection{Data Transfer Planning Context}

We consider a data transfer planning scenario where an AI system must generate and execute plans for transferring personal information across different jurisdictions and systems. Each plan $P$ consists of a sequence of actions $A = \{a_1, a_2, \ldots, a_n\}$, where each action $a_i$ involves the handling of personal information for a specific purpose.

\begin{definition}[Data Transfer Action]
A data transfer action $a_i$ is defined as a tuple $(c, p, d, o)$ where:
\begin{itemize}
\item $c$ represents the company context and data handling capabilities
\item $p$ represents the stated purpose of data utilization
\item $d$ represents the proposed data handling operation
\item $o$ represents additional operational context
\end{itemize}
\end{definition}

\subsection{APPI Article 16 Compliance}

APPI Article 16 imposes restrictions on handling personal information \emph{beyond the Purpose of Utilization} specified under Article 15, unless explicit prior consent is obtained from the data subject.

\begin{enumerate}
\item \textbf{Primary Rule (Paragraph 1)}: A business operator shall not handle personal information beyond the scope necessary for achieving the stated purpose, without prior consent of the individual.
\item \textbf{Business Succession Rule (Paragraph 2)}: If personal information is acquired through business succession (e.g., merger, acquisition), the same restriction applies with respect to the original purpose prior to the succession.
\item \textbf{Exceptions (Paragraph 3)}: The above restrictions do not apply when:
    \begin{enumerate}
    \item Handling is based on laws or regulations
    \item Necessary to protect life, body, or property where obtaining consent is difficult
    \item Specially necessary for public health or the sound growth of children where obtaining consent is difficult
    \item Necessary to cooperate with governmental authorities in legal duties where obtaining consent may impede execution
    \end{enumerate}
\end{enumerate}

\begin{definition}[Compliance Status]
\footnotesize
For a given action $a_i$, the compliance status $S(a_i) \in 
\{\text{COMPLIANT}, \text{NON-COMPLIANT}\}$ is determined by:
\[
S(a_i) =
\begin{cases}
\text{COMPLIANT}, & \text{if:} \\
& \quad \text{$d$ is within purpose $p$,} \\
& \quad \text{or prior consent is obtained,} \\
& \quad \text{or an exception applies} \\
\text{NON-COMPLIANT}, & \text{otherwise.}
\end{cases}
\]
\end{definition}

\subsection{Legal Verifier Problem}

The legal verifier problem can be formalized as a function $V: A \rightarrow S$ that maps each action to its compliance status. The challenge lies in accurately implementing this function given the complexity and ambiguity inherent in legal interpretation.

Traditional single-agent approaches attempt to implement $V$ directly through a monolithic reasoning system. Our multi-agent approach decomposes $V$ into specialized sub-functions:

\begin{align}
V(a_i) = C(L(a_i), X(a_i), R(a_i))
\end{align}

where:
\begin{itemize}
\item $L(a_i)$ represents legal analysis of the action
\item $X(a_i)$ represents contextual analysis of business necessity
\item $R(a_i)$ represents risk assessment and edge case evaluation
\item $C(\cdot)$ represents the coordination function that synthesizes these analyses
\end{itemize}

\section{Multi-Agent Legal Verifier Architecture}
\label{sec:system}


The section proposes a design of \textbf{multi-agent legal verifier system}, including agents and coordination protocols.

\subsection{Agent Specialization}

Our multi-agent legal verifier system consists of four specialized agents working in coordination to evaluate legal compliance.

\begin{enumerate}
\item \textbf{Legal Analyst Agent}: Specializes in interpreting legal requirements and precedents
\item \textbf{Context Analyzer Agent}: Focuses on business context and purpose alignment
\item \textbf{Risk Assessor Agent}: Evaluates privacy risks and edge cases
\item \textbf{Coordinator Agent}: Synthesizes analyses and makes final decisions
\end{enumerate}

\subsubsection{Legal Analyst Agent}

The Legal Analyst Agent ($L$) focuses on pure legal interpretation, analyzing whether proposed actions align with statutory requirements. Its primary responsibilities include:

\begin{itemize}
\item Interpreting the scope of "necessary for the achievement of the Purpose of Utilization"
\item Identifying when actions clearly fall within or outside legal boundaries
\item Applying legal precedents and interpretative guidelines
\end{itemize}

The agent operates with high confidence on clear legal violations but may express uncertainty on borderline cases, providing valuable signals to the coordination process.

\subsubsection{Context Analyzer Agent}

The Context Analyzer Agent ($X$) specializes in business context evaluation, determining whether proposed actions are reasonably necessary for stated business purposes. Key functions include:

\begin{itemize}
\item Evaluating business necessity and proportionality
\item Assessing alignment between stated purposes and proposed actions
\item Identifying legitimate business justifications for data handling
\end{itemize}

This agent provides crucial context that pure legal analysis might miss, particularly in complex business scenarios.

\subsubsection{Risk Assessor Agent}

The Risk Assessor Agent ($R$) focuses on privacy risk evaluation and edge case identification. Its responsibilities include:

\begin{itemize}
\item Assessing privacy risks associated with proposed actions
\item Evaluating consent mechanisms and their adequacy
\item Identifying special circumstances that might affect compliance
\item Analyzing potential consequences of data handling decisions
\end{itemize}

\subsubsection{Coordinator Agent}

The Coordinator Agent ($C$) synthesizes the analyses from all specialist agents to make final compliance determinations. It employs sophisticated reasoning to:

\begin{itemize}
\item Weight different perspectives based on confidence levels
\item Resolve conflicts between agent assessments
\item Apply meta-reasoning about legal uncertainty
\item Generate comprehensive justifications for decisions
\end{itemize}

\subsection{Coordination Protocol}

The multi-agent system follows a structured coordination protocol:


\begin{enumerate}
\item $analysis_L \leftarrow$ LegalAnalyst.analyze($a_i$)
\item $analysis_X \leftarrow$ ContextAnalyzer.analyze($a_i$)
\item $analysis_R \leftarrow$ RiskAssessor.analyze($a_i$)
\item $analyses \leftarrow \{analysis_L, analysis_X, analysis_R\}$
\item $(S(a_i), J(a_i)) \leftarrow$ Coordinator.synthesize($a_i$, $analyses$)
\end{enumerate}

\section{Experimental Design}
\label{sec:experiment}

This section describes the experiment design, including dataset construction, evaluation metrics, and baseline comparison. 

\subsection{Dataset Construction}

We constructed a comprehensive dataset of 200 test cases representing diverse compliance scenarios under APPI Article 16. The dataset is stratified across four categories:

\begin{itemize}
\item \textbf{Clear Compliance} (50 cases): Actions clearly within stated purposes
\item \textbf{Clear Violations} (50 cases): Actions clearly beyond stated purposes without consent
\item \textbf{Consent-Based Compliance} (50 cases): Actions beyond original purpose with explicit consent
\item \textbf{Edge Cases} (50 cases): Ambiguous scenarios requiring nuanced interpretation
\end{itemize}

\noindent Each test case includes:
\begin{itemize}
\item Company context and data handling capabilities
\item Stated purpose of data utilization
\item Proposed data handling action
\item Additional operational context
\item Ground truth compliance label with detailed reasoning
\end{itemize}

\subsection{Evaluation Metrics}

We evaluate system performance using standard classification metrics:

\begin{itemize}
\item \textbf{Accuracy}: Overall proportion of correct classifications
\item \textbf{Precision}: Proportion of predicted compliant cases that are actually compliant
\item \textbf{Recall}: Proportion of actually compliant cases correctly identified
\item \textbf{F1-Score}: Harmonic mean of precision and recall
\end{itemize}

Additionally, we analyze:
\begin{itemize}
\item \textbf{Category-specific performance}: Accuracy within each compliance category
\item \textbf{Confidence calibration}: Relationship between predicted confidence and actual accuracy
\item \textbf{Processing time}: Computational efficiency comparison
\end{itemize}

\subsection{Baseline Comparison}

We compare our multi-agent system against a single-agent baseline that attempts to perform all aspects of legal analysis within a unified reasoning framework. Both systems use identical underlying language models (GPT-3.5-turbo) to ensure fair comparison.

\section{Results and Analysis}
\label{sec:result}

This section reports the experiment results, including overall performance, category-specific analysis, confidence analysis, and processing time analysis.

\subsection{Overall Performance}

\begin{figure}[t]
\centering
\includegraphics[width=0.45\textwidth]{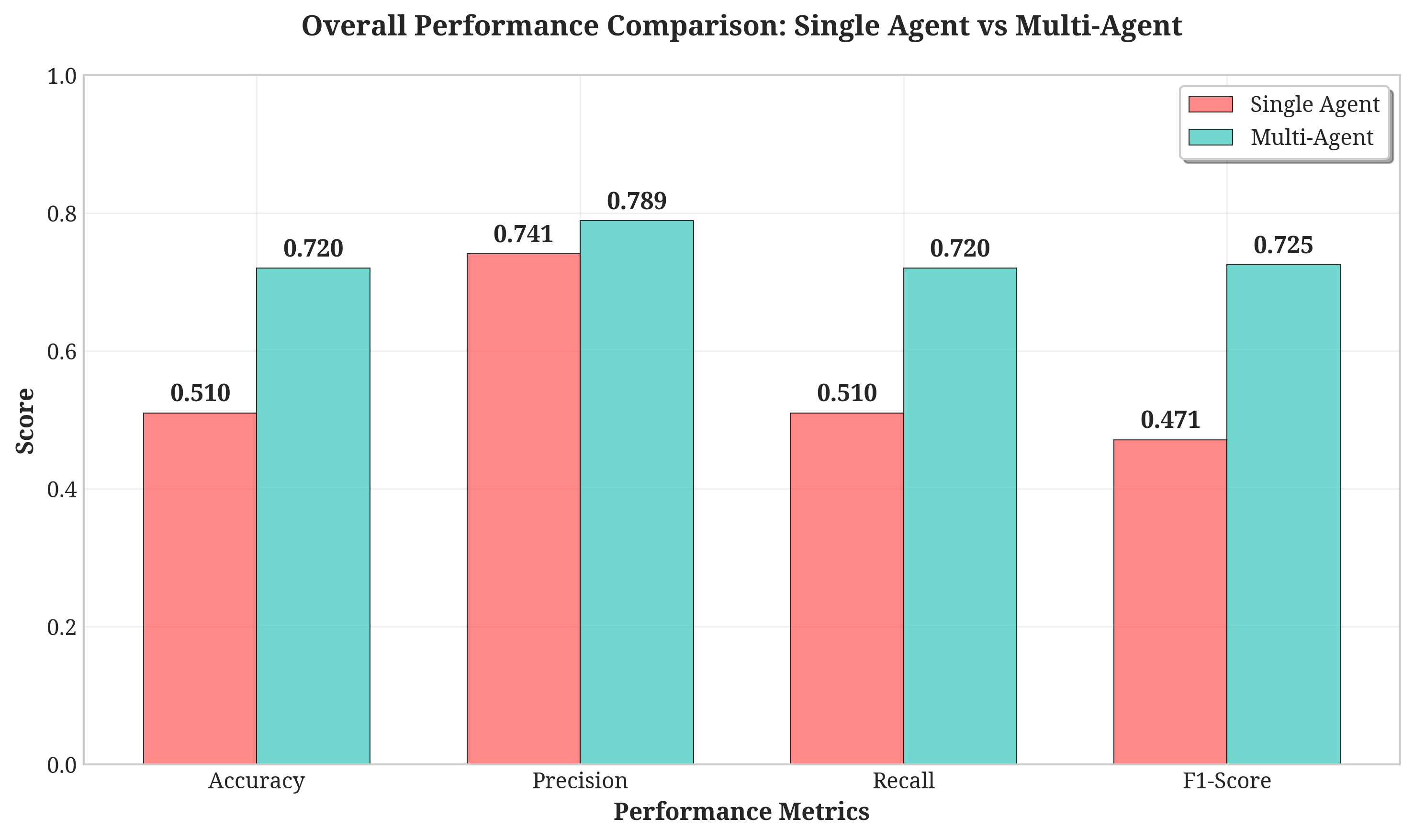}
\caption{Overall Performance Comparison between Single Agent and Multi-Agent approaches across all evaluation metrics. The multi-agent system shows substantial improvements in accuracy, recall, and F1-score.}
\label{fig:overall_performance}
\end{figure}

Table~\ref{tab:overall_performance} presents the comprehensive comparison between single-agent and multi-agent approaches. The visualization in Figure~\ref{fig:overall_performance} complements the tabular data, highlighting that the multi-agent system significantly outperforms the single-agent baseline across all key metrics.

\begin{table*}[t]
\centering
\caption{Overall Performance Comparison}
\label{tab:overall_performance}
\begin{tabular}{lcccc}
\toprule
\textbf{Approach} & \textbf{Accuracy} & \textbf{Precision} & \textbf{Recall} & \textbf{F1-Score} \\
\midrule
Single Agent & 0.510 & 0.741 & 0.510 & 0.471 \\
Multi-Agent & 0.720 & 0.789 & 0.720 & 0.725 \\
\midrule
\textbf{Improvement} & \textbf{+0.210} & \textbf{+0.048} & \textbf{+0.210} & \textbf{+0.255} \\
\textbf{(\% Change)} & \textbf{(+41.2\%)} & \textbf{(+6.5\%)} & \textbf{(+41.2\%)} & \textbf{(+54.1\%)} \\
\bottomrule
\end{tabular}
\end{table*}

The multi-agent system demonstrates substantial improvements across all metrics. Notably:
\begin{itemize}
    \item \textbf{Accuracy} shows a 41.2\% relative improvement, indicating better overall correctness of predictions.
    \item \textbf{F1-Score} gains 54.1\%, reflecting a balanced enhancement in both precision and recall.
    \item \textbf{Precision} improves modestly (+6.5\%), showing that the multi-agent approach retains the ability to avoid false positives.
    \item \textbf{Recall} matches the accuracy improvement, confirming that more relevant cases are correctly identified.
\end{itemize}
These results suggest that the multi-agent architecture enhances both the breadth (recall) and reliability (precision) of detection, with the greatest impact on balanced performance measures.

\subsection{Category-Specific Analysis}

\begin{figure*}[t]
\centering
\includegraphics[width=0.7\textwidth]{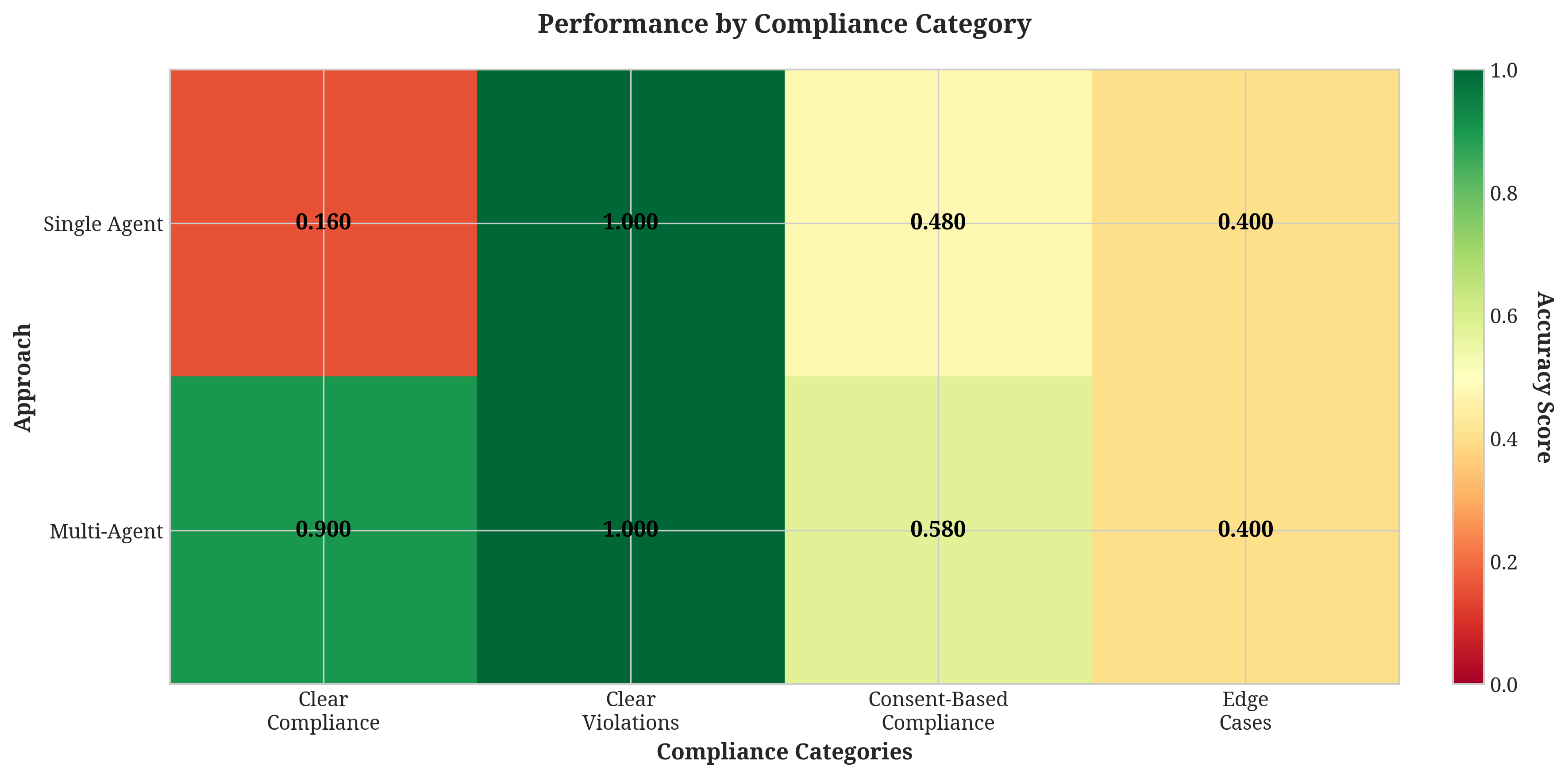}
\caption{Performance Heatmap by Compliance Category. The visualization clearly shows the multi-agent system's superior performance, particularly in Clear Compliance scenarios.}
\label{fig:category_heatmap}
\end{figure*}

Table~\ref{tab:category_performance} details the breakdown by compliance category, showing how performance gains vary depending on case characteristics.

\begin{table*}[t]
\centering
\caption{Performance by Compliance Category}
\label{tab:category_performance}
\begin{tabular}{lcccc}
\toprule
\textbf{Category} & \textbf{Single Agent} & \textbf{Multi-Agent} & \textbf{Improvement} & \textbf{Sample Count} \\
\midrule
Clear Compliance & 0.160 & 0.900 & +0.740 & 50 \\
Clear Violations & 1.000 & 1.000 & +0.000 & 50 \\
Consent-Based Compliance & 0.480 & 0.580 & +0.100 & 50 \\
Edge Cases & 0.400 & 0.400 & +0.000 & 50 \\
\bottomrule
\end{tabular}
\end{table*}

Analysis reveals:
\begin{itemize}
    \item \textbf{Clear Compliance}: Largest gain (+74.0 pp), suggesting that specialized agents are better at confirming legitimate, rule-abiding behavior.
    \item \textbf{Clear Violations}: Both approaches achieve perfect scores, implying that obvious breaches are trivial to detect regardless of architecture.
    \item \textbf{Consent-Based Compliance}: Moderate gain (+10.0 pp) points to improved handling of nuanced, consent-related decisions by leveraging specialized risk assessment.
    \item \textbf{Edge Cases}: No improvement observed, underscoring that truly ambiguous situations remain a challenge for both systems.
\end{itemize}
These patterns suggest that the multi-agent system delivers the most value in unambiguous compliance scenarios, with incremental benefits for moderately complex cases.

\subsection{Confidence Analysis}

\begin{figure*}
\centering
\includegraphics[width=0.7\textwidth]{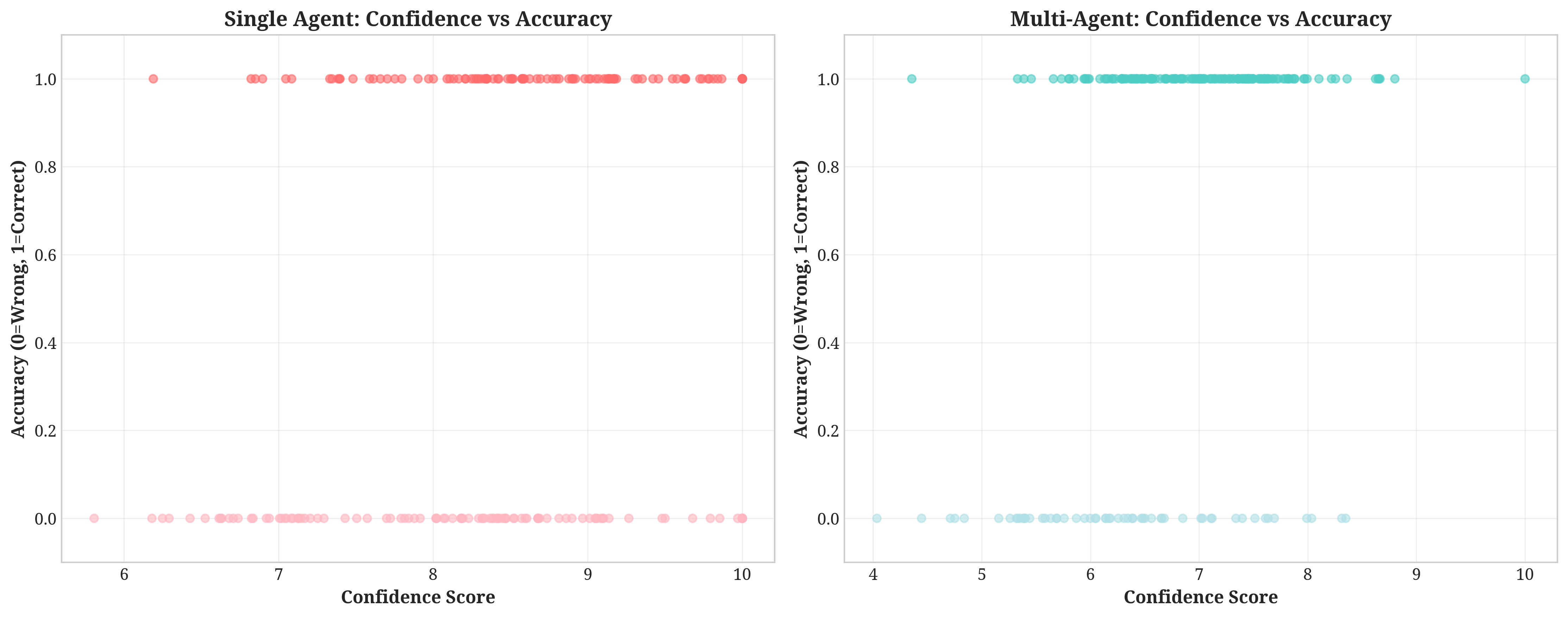}
\caption{Confidence vs Accuracy Analysis for Single Agent and Multi-Agent Systems. The multi-agent system shows better confidence calibration with more reliable confidence scores.}
\label{fig:confidence_analysis}
\end{figure*}

Figure~\ref{fig:confidence_analysis} compares how confidence scores map to actual prediction accuracy. Key findings:
\begin{itemize}
    \item The multi-agent system's confidence scores exhibit stronger correlation with true accuracy, indicating better calibration.
    \item High-confidence predictions from the multi-agent system are more trustworthy, reducing risk in high-stakes decisions.
    \item This improved calibration suggests that the system effectively synthesizes uncertainty estimates from specialized agents, resulting in a more informed and balanced final decision.
\end{itemize}
Such calibration improvements are crucial in operational contexts, where decision-makers rely on confidence scores to allocate review resources.

\subsection{Processing Time Analysis}

\begin{figure}
\centering
\includegraphics[width=0.45\textwidth]{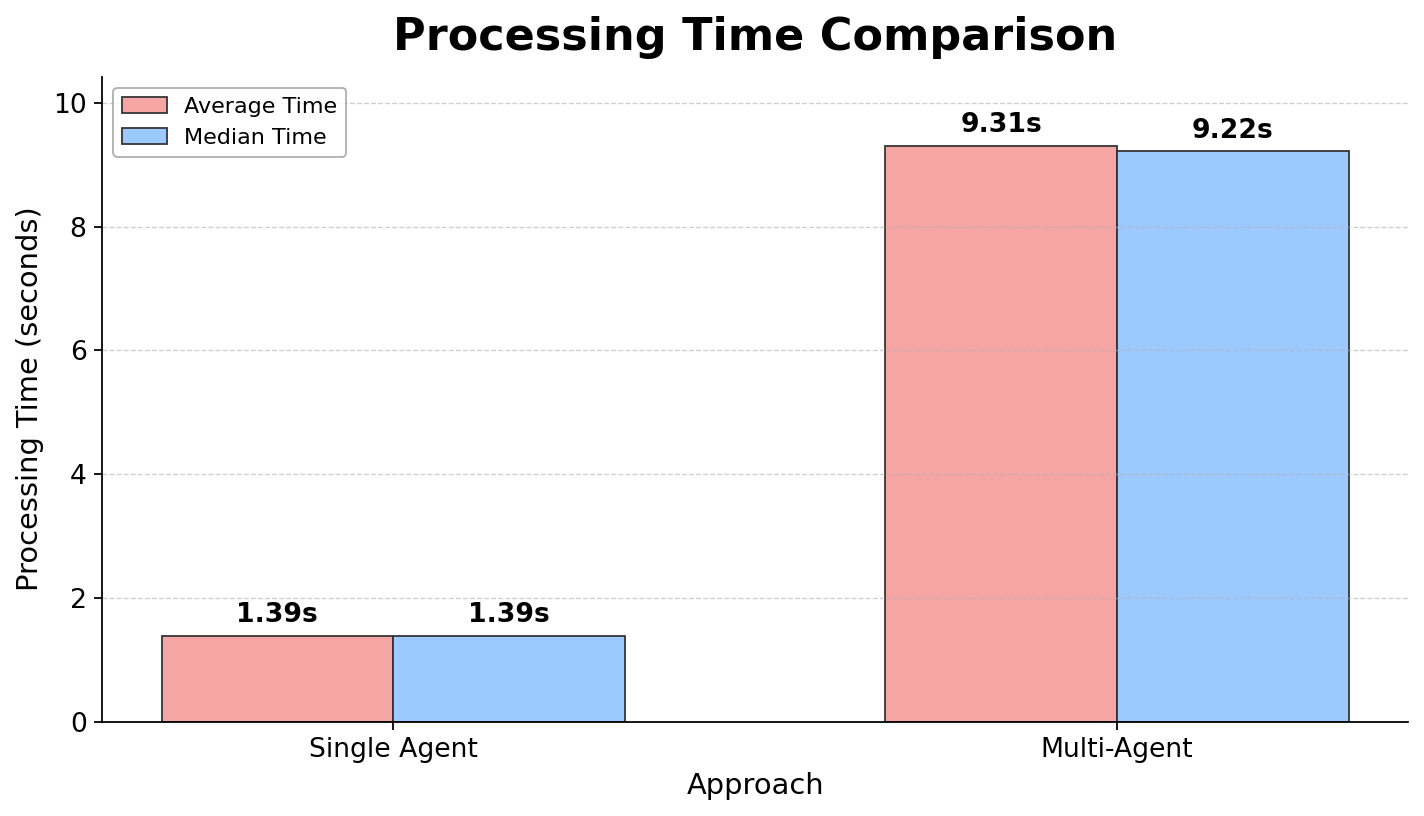}
\caption{Processing Time Comparison showing the computational overhead of the multi-agent approach. While the multi-agent system requires 6.67x more processing time, this may be acceptable for high-stakes compliance decisions.}
\label{fig:processing_time}
\end{figure}

The multi-agent approach incurs a 6.67x processing time overhead (9.31s vs 1.39s average), as shown in Figure~\ref{fig:processing_time}, reflecting the additional computational cost of specialized analysis and coordination. However, this overhead may be acceptable for high-stakes compliance decisions where accuracy is paramount.

\section{Discussion}
\label{sec:discussion}

This section discusses several findings of this paper, including its implications, architectural insights, limitations, and future work.

\subsection{Implications for Legal AI}

Our findings reinforce the view that legal reasoning—particularly in the domain of compliance assessment—is not a monolithic problem, but a layered cognitive task that benefits from distributed expertise. The substantial 21.0 percentage point increase in accuracy achieved by the multi-agent system represents not merely an incremental improvement, but a meaningful shift in the feasibility of automated legal compliance checking. 

This improvement carries direct implications for both the practice of law and the design of legal technology. In real-world deployments, such a system could serve as a front-line compliance filter, reducing the volume of manual reviews required by human legal teams. More importantly, the multi-agent architecture enables the AI to differentiate between superficially similar cases by applying context-sensitive reasoning, a capability that traditional single-agent models often lack. This is particularly critical in the regulatory space, where the same statutory provision may be interpreted differently depending on factual nuances, industry-specific norms, or jurisdictional precedents.

The category-specific performance analysis provides further clarity on the strengths of this approach. The exceptional improvement in Clear Compliance scenarios underscores the value of embedding domain-specific business context analysis directly into the reasoning process. By contrast, the more modest gains in Consent-Based Compliance cases suggest that some legal determinations remain inherently dependent on the subtleties of intent, documentation, and historical precedent—factors that cannot be fully captured by surface-level data analysis alone.

From a strategic perspective, these results hint at an emerging design principle for high-performance legal AI: instead of pursuing a single, universally competent model, it may be more effective to build a coalition of specialized reasoning modules, each optimized for a narrow domain of expertise and coordinated through a structured decision-making framework. Such systems are not only more interpretable, but also more resilient to the weaknesses of any individual reasoning pathway.

\subsection{Architectural Insights}

The success of our multi-agent architecture can be attributed to several key design decisions:

\begin{enumerate}
\item \textbf{Specialization}: Each agent focuses on a specific aspect of legal analysis, allowing for deeper expertise development and more targeted reasoning capabilities.
\item \textbf{Coordination}: The coordinator agent synthesizes diverse perspectives while managing uncertainty, ensuring that decisions reflect a balanced integration of viewpoints rather than a single dominant interpretation.
\item \textbf{Confidence Integration}: The system incorporates calibrated confidence scores from specialized agents, enabling it to weigh contributions proportionally and make more informed, risk-adjusted final decisions.
\end{enumerate}

\subsection{Limitations and Future Work}

Several limitations warrant discussion:

\begin{itemize}
\item \textbf{Edge Case Performance}: Neither approach shows improvement on genuinely ambiguous cases, suggesting fundamental challenges in legal uncertainty handling.
\item \textbf{Computational Cost}: The 6.67x processing overhead may limit practical deployment in time-sensitive applications.
\item \textbf{Evaluation Scope}: Our evaluation focuses on APPI Article 16; generalization to other regulations requires further investigation.
\end{itemize}

Building upon the identified limitations, future work will focus on enhancing the system's capabilities and practical applicability. A primary objective is to address the fundamental challenge of handling legal ambiguity. To move beyond the current performance ceiling on genuinely ambiguous cases, future iterations will integrate comprehensive legal precedent and case law databases. This will enable the AI to interpret statutes not merely as abstract rules, but in the context of their real-world application by courts, thereby fostering a more nuanced and sophisticated approach to legal reasoning in complex scenarios.

Furthermore, to broaden the system's utility beyond its current scope, we will explore dynamic agent specialization. Instead of being statically assigned to a single aspect of legal analysis, agents could adapt their expertise based on the specific regulatory domain in question, such as GDPR in Europe or CCPA in California. This dynamic specialization would not only test the model's generalizability but also pave the way for tackling complex multi-jurisdictional compliance scenarios, a critical need for global enterprises.

Finally, to mitigate the significant computational overhead that currently limits practical deployment, we will investigate hybrid approaches designed to balance accuracy and efficiency. For instance, the system could employ a simpler, faster model for an initial screening of straightforward cases, reserving the resource-intensive multi-agent architecture for situations that genuinely require deep, multifaceted analysis. Such a hybrid model would make the system viable for time-sensitive, real-world applications without substantially compromising the accuracy gains our research has demonstrated.

\section{Conclusion}
\label{sec:conclusion}

This paper presents a novel multi-agent legal verifier system for data transfer planning compliance, demonstrating significant improvements over single-agent approaches. Our experimental evaluation across 200 APPI Article 16 compliance scenarios shows a 21.0 percentage point accuracy improvement, with particularly strong performance in recognizing legitimate business uses of personal data.

The success of our approach validates the hypothesis that legal compliance checking benefits from specialized multi-agent architectures that decompose complex legal reasoning into manageable sub-problems. The coordination mechanism effectively synthesizes diverse analytical perspectives while maintaining appropriate uncertainty handling.

These results have important implications for the development of trustworthy AI systems operating under regulatory constraints. As privacy regulations continue to evolve and proliferate globally, automated compliance verification systems will become increasingly critical for AI deployment in data-sensitive applications.

Our work contributes to the growing intersection of AI and law by providing both a practical framework for automated compliance checking and empirical evidence supporting multi-agent approaches for complex legal reasoning tasks. The open questions raised by our edge case analysis point toward important directions for future research in legal AI systems.

\section*{Acknowledgement}

This work was partly supported by the Strategic Research Projects grant from ROIS, Research Organization of Information and Systems (2025-SRP-25).

\bibliographystyle{kr}
\bibliography{references}

\end{document}